\def\year{2020}\relax
\documentclass[letterpaper]{article} 
\usepackage{aaai20}  
\usepackage{times}  
\usepackage{helvet} 
\usepackage{courier}  
\usepackage[hyphens]{url}  
\usepackage{graphicx} 
\usepackage{graphicx}  
\usepackage{verbatim}
\usepackage[usenames,dvipsnames]{xcolor}
\usepackage{booktabs} 
\usepackage{pbox}
\usepackage{amsfonts}
\usepackage{amsmath}
\usepackage{multirow}

\newcommand{\ie}{\textit{i.e.}}
\newcommand{\eg}{\textit{e.g.}}

\newcommand{\stos}{seq2seq }
\newcommand{\stosns}{seq2seq} 
\newcommand{\bidir}{bidirectional }
\newcommand{\bidirns}{bidirectional}

\newcommand{\vlpns}{Vision-Language Pre-training}
\newcommand{\vlpabbr}{VLP }
\newcommand{\vlpabbrns}{VLP}
\newcommand{\cls}{\textsf{[CLS]} }
\newcommand{\sep}{\textsf{[SEP]} }
\newcommand{\stopc}{\textsf{[STOP]} }
\newcommand{\maskc}{\textsf{[MASK]} }
\newcommand{\clsns}{\textsf{[CLS]}}
\newcommand{\sepns}{\textsf{[SEP]}}
\newcommand{\stopcns}{\textsf{[STOP]}}

\newcommand{\softmax}{\textrm{softmax}}

\newcommand{\head}[1]{\noindent\textbf{#1}}

\title{Unified Vision-Language Pre-Training for Image Captioning and VQA}

\author{Luowei Zhou\textsuperscript{\rm 1}, \Large \textbf{Hamid Palangi\textsuperscript{\rm 2}, Lei Zhang\textsuperscript{\rm 3},}  \Large \textbf{Houdong Hu\textsuperscript{\rm 4},} \Large \textbf{Jason J. Corso\textsuperscript{\rm 1}, Jianfeng Gao\textsuperscript{\rm 2}} \\ 
\textsuperscript{\rm 1} University of Michigan \;\;
\textsuperscript{\rm 2} Microsoft Research \;\;
\textsuperscript{\rm 3} Microsoft Cloud \& AI \;\;
\textsuperscript{\rm 4} Microsoft AI \& Research \\
{\{luozhou, jjcorso\}@umich.edu \;\;\; \{hpalangi, leizhang, houhu, jfgao\}@microsoft.com} 
}

\begin{document}

\maketitle

\begin{abstract}
This paper presents a unified Vision-Language Pre-training (VLP) model. The model is \emph{unified} in that (1) it can be fine-tuned for either vision-language generation (\eg, image captioning) or understanding (\eg, visual question answering) tasks, and (2) it uses a shared multi-layer transformer network for both encoding and decoding, which differs from many existing methods where the encoder and decoder are implemented using separate models. The unified VLP model is pre-trained on a large amount of image-text pairs using the unsupervised learning objectives of two tasks: bidirectional and sequence-to-sequence (seq2seq) masked vision-language prediction. The two tasks differ solely in what context the prediction conditions on. This is controlled by utilizing specific self-attention masks for the shared transformer network. To the best of our knowledge, \vlpabbr is the first reported model that achieves state-of-the-art results on both vision-language generation and understanding tasks, as disparate as image captioning and visual question answering, across three challenging benchmark datasets: COCO Captions, Flickr30k Captions, and VQA 2.0. The code and the pre-trained models are available at \url{https://github.com/LuoweiZhou/VLP}.
\end{abstract}

\section{Introduction}\label{sec:intro}

Inspired by the recent success of pre-trained language models such as BERT~\cite{devlin2018bert} and GPT~\cite{radford2018improving,radford2019language}, there is a growing interest in extending these models to learning cross-modal representations like image-text~\cite{lu2019vilbert,tan2019lxmert} and video-text~\cite{sun2019videobert,sun2019contrastive}, 
for various vision-language tasks such as Visual Question Answering (VQA) and video captioning, where traditionally tedious task-specific feature designs and fine-tuning are required.

\begin{figure}[!t]
    \centering
    \includegraphics[width=\linewidth]{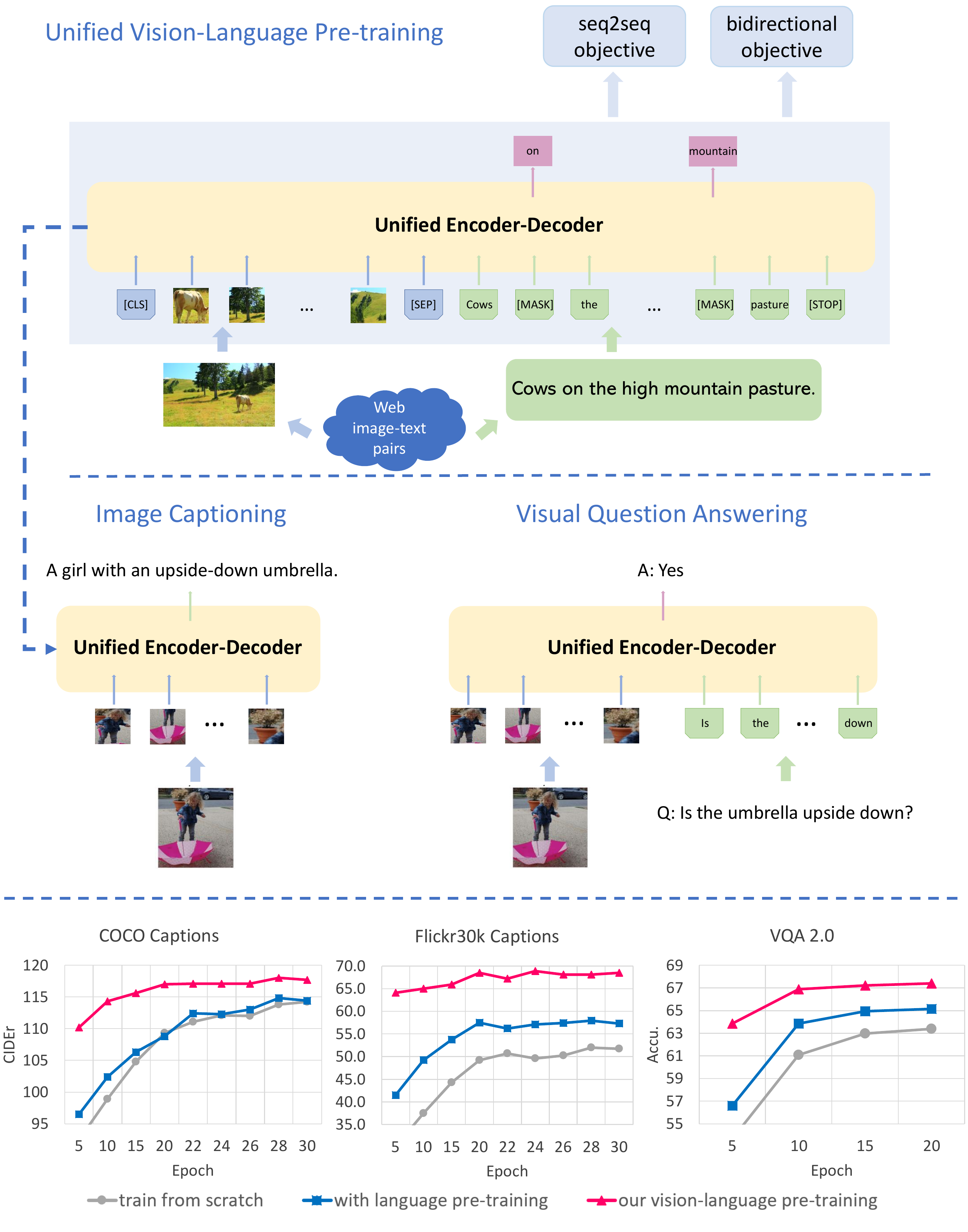}
    \caption{We propose a unified encoder-decoder model for general vision-language pre-training. The pre-trained model is then fine-tuned for image captioning and visual question answering. Thanks to our vision-language pre-training, both training speed and overall accuracy have been significantly improved on the downstream tasks compared to random initialization or language-only pre-training. All the results are evaluated on the validation set of the corresponding dataset.} 
    \label{fig:teaser}
\end{figure}

\begin{table*}[t]
    \centering
    \scriptsize{ 
    \begin{tabular}{l|l|l|l|l}
        \toprule
        Type & Method & Domain & Architecture  & Downstream Tasks  \\
        \midrule
        \pbox{10cm}{Understanding-based only} & \pbox{10cm}{LXMERT~\cite{tan2019lxmert},\\ ViLBERT~\cite{lu2019vilbert}, UNITER~\cite{chen2019uniter},\\VisualBERT~\cite{li2019visualbert}, B2T2~\cite{alberti2019fusion}, \\Unicoder-VL~\cite{li2019unicoder},VL-BERT~\cite{su2019vl}} & Image & \pbox{10cm}{Single-stream or \\two stream Transformer} & \pbox{10cm}{Visual question answering\\Visual commonsense reasoning\\Image retrieval\\Grounding referring expressions}  \\
        \midrule[0.08em]
        \multirow{6}{*}{\pbox{10cm}{Generation-based and \\ understanding-based}} & VideoBERT~\cite{sun2019videobert} & Video & \pbox{10cm}{Single-stream Transformer+\\Masked Transformer~\cite{zhou2018end}}   & \pbox{10cm}{Zero-shot action classification \\ Video captioning} \\
        \cmidrule{2-5}
        & CBT~\cite{sun2019contrastive} & Video & \pbox{10cm}{Two-stream Transformer encoder+\\Transformer decoder}  & \pbox{10cm}{Action anticipation \\ Video captioning} \\      
        \cmidrule[0.08em]{2-5}
        & Our \vlpabbr & Image & Single unified encoder-decoder  & \pbox{10cm}{Visual question answering \\ Image captioning }  \\
        \bottomrule
    \end{tabular}
    }
    \caption{Comparison between our method and other vision-language pre-training works.}
    \label{tab:related_work}
\end{table*}

Table~\ref{tab:related_work} summarizes some of the recent works on vision-language pre-training where all the models are unexceptionally built upon Bidirectional Encoder Representations from Transformers (BERT)~\cite{devlin2018bert}. These models use a two-stage training scheme. The first stage, called pre-training, learns the contextualized vision-language representations by predicting the masked words or image regions based on their intra-modality or cross-modality relationships 
on large amounts of image-text pairs.
Then, in the second stage, the pre-trained model is fine-tuned to adapt to a downstream task.

Although significant improvements have been reported on individual downstream tasks using different pre-trained models, it remains challenging to pre-train a \emph{single, unified} model that is universally applicable, via fine-tuning, to a wide range of vision-language tasks as disparate as vision-language generation (\eg, image captioning) and understanding (\eg, VQA).
Most existing pre-trained models are either developed only for understanding tasks, as denoted by ``understanding-based only'' in Tab.~\ref{tab:related_work}, or designed as  hybrid models that consist of multiple modality-specific encoders and decoders which have to be trained separately in order to support generation tasks.
For example, VideoBERT and CBT in Tab.~\ref{tab:related_work} 
perform pre-training only for the encoder, not for the decoder.
This causes a discrepancy between the cross-modal representations learned by the encoder and the representation needed by the decoder for generation, which could hurt the generality of the model. 
In this paper, we strive to develop a new method of pre-training a unified representation for both encoding and decoding, eliminating the aforementioned discrepancy. 
In addition, we expect that such a unified representation would also allow more effective cross-task knowledge sharing, reducing the development cost by eliminating the need of pre-training different models for different types of tasks.

To this end, we propose a unified encoder-decoder model, called the Vision-Language Pre-training (VLP) model, which can be fine-tuned for both vision-language generation
and understanding
tasks.
The VLP model uses a shared multi-layer Transformer network~\cite{vaswani2017attention} for encoding and decoding, pre-trained on large amounts of image-caption pairs~\cite{sharma2018conceptual}, and optimized for two unsupervised vision-language prediction tasks: bidirectional and sequence to sequence (seq2seq) masked language prediction.
The two tasks differ solely in what context the prediction conditions on.
This is controlled by utilizing specific self-attention masks for the shared Transformer network.
In the \bidir prediction task, the context of the masked caption word to be predicted consists of all the image regions and all the words on its right and left in the caption. 
In the \stos task, the context consists of all the image regions and the words on the left of the to-be-predicted word in the caption. 

The proposed \vlpabbr has two main advantages in comparison with the BERT-based models in Tab.~\ref{tab:related_work}. 
First, VLP unifies the encoder and decoder and learns a more universal contextualized vision-language representation that can be more easily fine-tuned for vision-language generation and understanding tasks, as disparate as image captioning and VQA. 
Second, the unified pre-training procedure leads to a single model architecture for two distinct vision-language prediction tasks, \ie, bidirectional and seq2seq, alleviating the need for multiple pre-training models for different types of tasks without any significant performance loss in task-specific metrics.

We validate VLP in our experiments on both the image captioning and VQA tasks using three challenging benchmarks: COCO Captions~\cite{chen2015microsoft}, Flickr30k Captions~\cite{young2014image}, and VQA 2.0 dataset~\cite{goyal2017making}.
We observe that compared to the two cases where we do not use any pre-trained model or use only the pre-trained language model (\ie, BERT), using VLP significantly speed-ups the task-specific fine-tuning and leads to better task-specific models,
as shown in Fig.~\ref{fig:teaser}. 
More importantly, without any bells and whistles, our models 
achieve state-of-the-art results on both tasks across all three datasets.

\section{Related Work}\label{sec:related_work}
\head{Language Pre-training.} Among numerous BERT variants in language pre-training, we review the two methods that are most relevant to our approach, namely Unified LM or UniLM~\cite{dong2019unified} and Multi-Task DNN (MT-DNN)~\cite{liu2019multi}. UniLM employs a shared Transformer network which is pre-trained on three language modeling objectives: unidirectional, bidirectional, and sequence-to-sequence. Each objective specifies different binary values in the self-attention mask to control what context is available to the language model. MT-DNN combines multi-task training and pre-training by attaching task-specific projection heads to the BERT network. Our work is inspired by these works and tailored for vision-language tasks in particular.

\head{Vision-Language Pre-training.} This has become a nascent research area in the vision-language community. Related works include ViLBERT~\cite{lu2019vilbert} and LXMERT~\cite{tan2019lxmert}, both of which tackle understanding-based tasks only (e.g., VQA and Retrieval) and share the same two-stream BERT framework with a vision-language co-attention module to fuse the information from both modalities. ViLBERT is tested on a variety of downstream tasks including VQA, referring expression, and image-to-text retrieval. LXMERT only focuses on a particular problem space (\ie, VQA and visual reasoning) and the generalization ability further compromises when the datasets from the downstream tasks are also exploited in the pre-training stage. 
The most similar work to ours is VideoBERT~\cite{sun2019videobert}, which addresses generation-based tasks (\eg, video captioning) and understanding-based tasks (\eg, action classification). However, it separates the visual encoder and the language decoder and performs pre-training only on the encoder, leaving decoder uninitialized.
In contrast, we propose a unified model for both encoding and decoding and fully leverage the benefit of pre-training. 

\head{Image Captioning \& VQA.}
Most of the recent works on image captioning are built upon~\cite{anderson2018bottom}, where a language model gets clues for sentence generation through dynamically attending on object regions in the image extracted from pre-trained object detectors.
Follow-up works further capture the relationships among object regions by using Graph Convolutional Networks (GCNs)~\cite{yao2018exploring}, incorporating language inductive bias~\cite{yang2019auto}, or enforcing region grounding between image and text~\cite{lu2018neural,Zhou_2019_CVPR}.
VQA is another prevalent research area in vision and language. Since its initial proposal~\cite{VQA0}, there has been a significant amount of works proposing model architectures to fuse question and image representations~\cite{kim2018bilinear,anderson2018bottom,gao2019dynamic}, 
new datasets or models to reduce the dataset bias~\cite{balanced_binary_vqa,goyal2017making,Agrawal2017DontJA} and ground the answer in the question~\cite{lewis2018generative}. We use our base architecture to perform both image captioning and VQA with minor model structure differences.

\section{Vision-Language Pre-training}\label{sec:VLP}

We denote the input image as $I$ and the associated/target sentence description (words) as $S$. We extract a fixed number $N$ of object regions from the image using an off-the-shelf object detector, denoted as $\{r_1,\ldots, r_N\}$ and the corresponding region features as $R=[R_1,\ldots, R_N]\in \mathbb{R}^{d\times N}$, region object labels (probabilities) as $C=[C_1,\ldots, C_N]\in \mathbb{R}^{l\times N}$, and region geometric information as $G=[G_1,\ldots, G_N]\in \mathbb{R}^{o\times N}$, where $d$ is the embedding size, $l$ indicates the number of the object classes of the object detector, and $o=5$ consists of four values for top left and bottom right corner coordinates of the region bounding box (normalized between 0 and 1) and one value for its relative area (\ie, ratio of the bounding box area to the image area, also between 0 and 1). The words in
$S$ are represented as one-hot vectors which are further encoded to word embeddings with embedding size $e$: $y_t \in \mathbb{R}^e$ where $t \in \{1, 2,\ldots, T\}$ and $T$ indicates the length of the sentence.

\begin{figure*}[t]
    \centering
    \includegraphics[width=\linewidth]{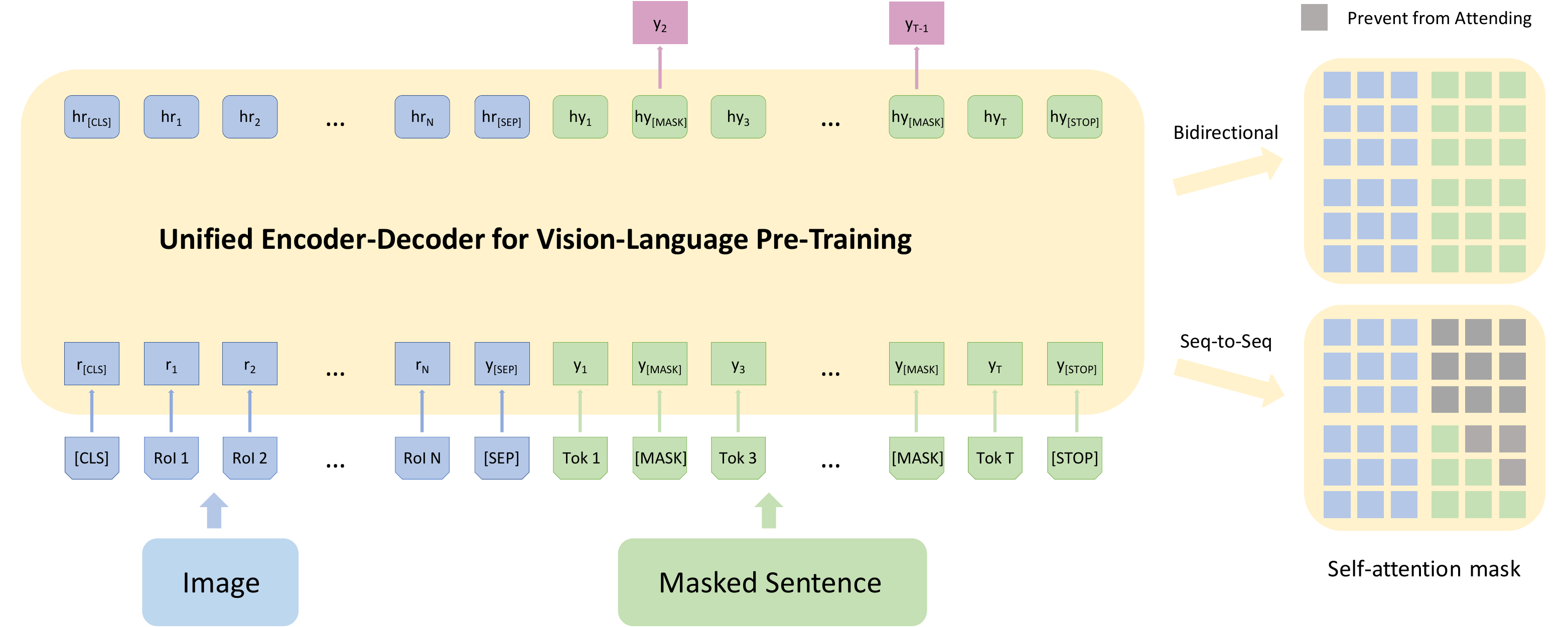}
    \caption{Model architecture for pre-training. The input comprises of image input, sentence input, and three special tokens (\clsns, \sepns, \stopcns). The image is processed as $N$ Region of Interests (RoIs) and region features are extracted according to Eq.~\ref{eq:region_feat}. The sentence is tokenized and masked with \maskc tokens for the later masked language modeling task. Our Unified Encoder-Decoder consists of 12 layers of Transformer blocks, each having a masked self-attention layer and feed-forward module, where the self-attention mask controls what input context the prediction conditions on. We implemented two self-attention masks depending on whether the objective is \bidir or \stosns. Better viewed in color.}
    \label{fig:model_arch}
\end{figure*}

\subsection{Vision-Language Transformer Network}
Our vision-language Transformer network, which unifies the Transformer encoder and decoder into a single model, is depicted in Fig.~\ref{fig:model_arch} (left). The model input consists of the class-aware region embedding, word embedding and three special tokens. The region embedding is defined as:
\begin{equation}\label{eq:region_feat}
    r_i = W_rR_i+W_p[\textrm{LayerNorm}(W_cC_i)|\textrm{LayerNorm}(W_gG_i)]
\end{equation}
where $[\cdot|\cdot]$ indicates the concatenation on the feature dimension, LayerNorm represents Layer Normalization. 
The second term mimics the positional embedding in BERT, but adding extra region class information, and $W_r, W_p, W_c, W_g$ are the embedding weights (the bias term and the nonlinearity term are omitted). 
Note that here we overload the notation of $r_i\in \mathbb{R}^d\;(i \in \{1, 2, . . . , N\})$ to also represent class-aware region embeddings.
In addition, we add segment embeddings to $r_i$ as in BERT where all the regions share the same segment embedding where the values depend on the objectives (\ie, \stos and \bidirns, see the following section).

The word embeddings are similarly defined as in~\cite{devlin2018bert}, adding up $y_t$ with positional embeddings and segment embeddings, which is again overloaded as $y_t$. We define three special tokens \clsns, \sepns, \stopcns, where \cls indicates the start of the visual input, \sep marks the boundary between the visual input and the sentence input, and \stopc determines the end of the sentence. The \maskc tokens indicate the masked words which will be explained in the next section.

\subsection{Pre-training Objectives}
In the BERT masked language modeling objective, 15\% of the input text tokens are first replaced with either a special \maskc token, a random token or the original token, at random with chances equal to 80\%, 10\%, and 10\%, respectively. Then, at the model output, the hidden state from the last Transformer block is projected to word likelihoods where the masked tokens are predicted in the form of a classification problem. Through this reconstruction, the model learns the dependencies in the context and forms a language model. We follow the same scheme and consider two specific objectives: 
the bidirectional objective (bidirectional) as in BERT and the sequence to sequence objective (seq2seq), inspired by~\cite{dong2019unified}.

As shown in Fig.~\ref{fig:model_arch} (right), the only difference between the two objectives lie in the self-attention mask. The mask used for the \bidir objective allows unrestricted message passing between the visual modality and the language modality while in \stosns, the to-be-predicted word cannot attend to the words in the future, \ie, it satisfies the auto-regressive property. More formally, we define the input to the first Transformer block as $H^0=[r_{\cls}, r_1,\ldots, r_N, y_{\sep}, y_1, \ldots, y_T, y_{\stopc}]\in \mathbb{R}^{d\times U}$ where $U=N+T+3$, and then the encoding at different levels of Transformer as $H^l=\textrm{Transformer}(H^{l-1}),\; l\in [1, L]$. We further define a self-attention mask as $M\in \mathbb{R}^{U\times U)}$, where
\begin{equation}
    M_{jk} =
    \begin{cases}
        0, \;\; \textrm{allow to attend} \\
        -\infty, \;\; \textrm{prevent from attending}
    \end{cases} j,k=1,\ldots, U.
\end{equation}
For simplicity, we assume a single attention head in the self-attention module. Then, the self-attention output on $H^{l-1}$ can be formulated as:
\begin{align}
    A^l= & \softmax(\frac{Q^\top K}{\sqrt{d}}+M)V^\top, \\
    V= & W_V^lH^{l-1}, \; Q=W_Q^lH^{l-1}, \; K=W_K^lH^{l-1},
\end{align}
\noindent where $W_V^l$, $W_Q^l$, and $W_K^l$ are the embedding weights (the bias terms are omitted). The intermediate variables $V$, $Q$, and $K$ indicate values, queries and keys, respectively, as in the self-attention module~\cite{vaswani2017attention}. $A^l$ is further encoded by a feed-forward layer with a residual connection to form the output $H^l$. During the pre-training, we alternate per-batch between the two objectives and the proportions of \stos and \bidir are determined by hyper-parameters $\lambda$ and $1-\lambda$, respectively.

It is worth noting that in our experiments we find that incorporating the region class probabilities ($C_i$) into region feature ($r_i$) leads to better performance than having a masked region classification pretext as in ~\cite{lu2019vilbert,tan2019lxmert}. 
Therefore, differing from existing works where masked region prediction tasks are used to refine the visual representation, we indirectly refine the visual representation by utilizing it for masked language reconstruction.
We also choose not to use the Next Sentence Prediction task as in BERT, or in our context predicting the correspondence between image and text,
because the task is not only weaker than \stos or \bidir but also computationally expensive. 
This coincidentally agrees with a concurrent work of RoBERTa~\cite{liu2019roberta}.

\head{Sequence-to-sequence inference.} Similar to the way \stos training is performed, we can directly apply VLP to sequence-to-sequence inference, in the form of beam search. More details follow next in the Image Captioning section.

\section{Fine-Tuning for Downstream Tasks}\label{sec:FT}
\subsection{Image Captioning}
We fine-tune the pre-trained \vlpabbr model on the target dataset using the \stos objective. During inference, we first encode the image regions along with the special \cls and \sep tokens and then start the generation by feeding in a \maskc token and sampling a word from the word likelihood output (\eg, greedy sampling). Then, the \maskc token in the previous input sequence is replaced by the sampled word and a new \maskc token is appended to the input sequence to trigger the next prediction.
The generation terminates when the \stopc token is chosen. Other inference approaches like beam search could apply as well.

\subsection{Visual Question Answering}
We frame VQA as a multi-label classification problem.
In this work we focus on open domain VQA where top $k$ most frequent answers are selected as answer vocabulary and used as class labels. Following~\cite{anderson2018bottom} we set $k$ to $3129$.

\begin{table*}[t]
    \centering
    {\small
    \begin{tabular}{l|rrrr|rrrr|rrrr} 
        \toprule
        & \multicolumn{4}{c}{COCO}  & \multicolumn{4}{c}{VQA 2.0 (Test-Standard)} & \multicolumn{4}{c}{Flickr30k} \\
        Method & B@4 & M & C & S & Overall & Yes/No & Number & Other & B@4 & M & C & S \\
        \midrule
        BUTD~\cite{anderson2018bottom} & 36.2 & 27.0 & 113.5 & 20.3 & 65.7 & - & - & - & 27.3 & 21.7 & 56.6 & 16.0  \\
        NBT (with BBox)~\cite{lu2018neural} & 34.7 & 27.1 & 107.2 & 20.1 & - & - & - & - & 27.1 & 21.7 & 57.5 & 15.6 \\
        GCN-LSTM (spa)~\cite{yao2018exploring} & \textbf{36.5} & 27.8 & 115.6 & 20.8 & - & - & - & - & - & - & - & - \\
        GCN-LSTM (sem) & \textbf{36.8} & 27.9 & 116.3 & 20.9 & - & - & - & - & - & - & - & - \\
        GVD~\cite{Zhou_2019_CVPR} & - & - & - & - & - & - & - & - & 26.9 & 22.1 & 60.1 & 16.1  \\
        GVD (with BBox) & - & - & - & - & - & - & - & - &  27.3 & 22.5 & 62.3 & 16.5 \\
        BAN~\cite{kim2018bilinear} & - & - & - & - & 70.4 & 85.8 & \textbf{53.7} & \textbf{60.7} & - & - & - & - \\ 
        DFAF~\cite{gao2019dynamic} & - & - & - & - & 70.3 & - & - & - & - & - & - & -  \\
        \midrule
        AoANet*~\cite{huang2019attention} & 37.2 & 28.4 & 119.8 & 21.3 & - & - & - & - & - & - & - & - \\
        ViLBERT*~\cite{lu2019vilbert} & - & - & - & - & 70.9 & - & - & - & - & - & - & - \\
        LXMERT*~\cite{tan2019lxmert} & - & - & - & - & 72.5 & 88.2 & 54.2 & 63.1 & - & - & - & - \\
        \midrule
        \textbf{Ours} \\
        \;\;w/o \vlpabbr pre-training (baseline) & 35.5 & 28.2 & 114.3 & 21.0 & 70.0 & 86.3 & 52.2 & 59.9 & 27.6 & 20.9 & 56.8 & 15.3 \\
        \;\;\stos pre-training only & \textbf{36.5} & \textbf{28.4} & \textbf{117.7} & \textbf{21.3} & 70.2 & 86.7 & 52.7 & 59.9 & \textbf{31.1} & \textbf{23.0} & \textbf{68.5} & \textbf{17.2} \\
        \;\;\bidir pre-training only & 36.1 & 28.3 & 116.5 & 21.2 & \textbf{71.3} & \textbf{87.6} & \textbf{53.5} & \textbf{61.2} & \textbf{30.5} & 22.6 & 63.3 & 16.9 \\
        \;\;Unified \vlpabbr & \textbf{36.5} & \textbf{28.4} & \textbf{116.9} & \textbf{21.2} & \textbf{70.7} & \textbf{87.4} & 52.1 & 60.5 & 30.1 & \textbf{23.0} & \textbf{67.4} & \textbf{17.0}  \\
        \bottomrule
    \end{tabular}
    \caption{Results on COCO Captions test set (with cross-entropy optimization only, all single models), VQA 2.0 Test-Standard set and Flickr30k test set. \textbf{* indicates unpublished works}. B@4 represents for BLEU@4, M for METEOR, C for CIDEr, and S for SPICE. Results on previous works are obtained from the original papers. Top two results on each metric are in bold.}
    \label{tab:test_results}
    }
\end{table*}

\begin{table}[t]
    \centering
    {\small
    \begin{tabular}{l|rrrr}
    \toprule
         & \multicolumn{4}{c}{COCO (w/ CIDEr optimization)} \\
        Method & B@4 & M & C & S \\
        \midrule
        BUTD & 36.3 & 27.7 & 120.1 & 21.4 \\
        GCN-LSTM (spa) & 38.2 & 28.5 & 127.6 & 22.0 \\
        SGAE~\cite{yang2019auto} & 38.4 & 28.4 & 127.8 & 22.1 \\
        \midrule
        AoANet* & 38.9 & 29.2 & 129.8 & 22.4 \\
        \midrule
        \textbf{Ours (Unified VLP)} & \textbf{39.5} & \textbf{29.3} & \textbf{129.3} & \textbf{23.2} \\
    \bottomrule    
    \end{tabular}
    }
    \caption{Results on COCO Captions test set (with CIDEr optimization, all single models). \textbf{* indicates unpublished works}. Top one result on each metric is in bold.}
    \label{tab:cider_optim}
\end{table}

During the fine-tuning, a multi-layer Perceptron (Linear+ReLU+Linear+Sigmoid) on top of the element-wise product of the last hidden states of \cls and \sep is learned, similar to~\cite{lu2019vilbert}.
We optimize the model output scores with respect to the soft answer labels using cross-entropy loss.
Note that unlike~\cite{tan2019lxmert}
where the task-specific objective (\ie, VQA) is exploited during pre-training by using the target datasets (from intensive human annotations), our pre-training does not have this requirement and is therefore more general.

\begin{table*}[t]
    \centering
    {\small
    \begin{tabular}{l|rrrr|rrrr|rrrr} 
        \toprule
        & \multicolumn{4}{c}{COCO}  & \multicolumn{4}{c}{VQA 2.0 (Test-Dev)} & \multicolumn{4}{c}{Flickr30k} \\
        Method & B@4 & M & C & S & Overall & Yes/No & Number & Other & B@4 & M & C & S \\
        \midrule
        From scratch & 35.2 & 27.9 & 112.5 & 20.6 & 67.7 & 83.5 & 50.7 & 58.1 & 28.4 & 20.8 & 53.5 & 15.2 \\
        Init from BERT & 34.8 & 28.1 & 112.6 & 20.7 & 68.6 & 85.2 & 50.9 & 58.3 & 29.1 & 21.7 & 60.4 & 15.9   \\
        Init from UniLM & 35.5 & 28.2 & 114.3 & 21.0 & 69.6 & 86.1 & \textbf{52.4} & 59.4 & 27.6 & 20.9 & 56.8 & 15.3 \\
        Unified \vlpabbr & \textbf{36.5} & \textbf{28.4} & \textbf{116.9} & \textbf{21.2} & \textbf{70.5} & \textbf{87.2} & 52.1 & \textbf{60.3} & \textbf{30.1} & \textbf{23.0} & \textbf{67.4} & \textbf{17.0}   \\
        \bottomrule
    \end{tabular}
    }
    \caption{Impact of different levels of pre-training on downstream tasks. All results are on the test set (Test-Dev for VQA 2.0). Top one result on each metric is in bold.}
    \label{tab:results_pt_levels}
\end{table*}

\begin{table}[t]
    \centering
    {\small
    \begin{tabular}{l|rrrr}
        \toprule
        Method & B@4 & M & C & S \\
        \midrule
        From scratch & 5.5 & 9.4 & 63.8 & 14.9 \\ 
        Init from BERT & 5.7 & 9.7 & 66.7 & 15.3 \\ 
        Init from UniLM & 5.8 & 9.7 & 67.0 & 15.5 \\
        \bottomrule
    \end{tabular}
    }
    \caption{Impact of model weight initializations on pre-training. Results are on Conceptual Captions val set on caption generation.} 
    \label{tab:cc_results}
\end{table}

\section{Experiments and Results}
\label{sec:exp}
\head{Data preparation.} We conduct pre-training on the Conceptual Captions (CC) dataset~\cite{sharma2018conceptual} which has around 3 million web-accessible images with associated captions. The datasets for downstream tasks include COCO Captions~\cite{chen2015microsoft}, VQA 2.0~\cite{goyal2017making} and Flickr30k~\cite{young2014image}. For COCO Captions and Flickr30k, we follow Karpathy's split\footnote{cs.stanford.edu/people/karpathy/deepimagesent/caption\_datasets.zip}, which gives 113.2k/5k/5k and 29.8k/1k/1k images for train/val/test splits respectively. For VQA 2.0, we split the dataset with the official partition, \ie, 443.8k questions from 82.8k images for training, 214.4k questions from 40.5k images for validation and report the results on Test-Standard set through the official evaluation server. We trim long sentences and pad short sentences to 20 words and all the words are tokenized and numericalized as in BERT~\cite{devlin2018bert}.

\head{Implementation details.}
Our Transformer backbone is the same as BERT-base~\cite{devlin2018bert}. The input of the network consists of image (regions) and the associated/target caption.
We represent each input image as 100 object regions extracted from a variant of Faster R-CNN~\cite{ren2015faster} pre-trained on Visual Genome~\cite{krishna2017visual,anderson2018bottom}. 
We take the model output from fc6 layer as the region feature ($R_i$) and the class likelihood on the 1600 object categories as region object labels ($C_i$). 
Note that if not specified, the weights in our BERT model are initialized from UniLM~\cite{dong2019unified} pre-trained on text corpora only.
For caption inference, we use greedy search on the validation set and beam search with beam size 5 on the test set.
We perform light model hyper-parameter search with the configurations presented in Appendix. 
$\lambda$ is set to 0.75 for CC pre-training from light model validation (out of $\{0.25, 0.5, 0.75\}$), and set to 1 for image captioning (\ie, full \stosns) and 0 for VQA (\ie, full \bidirns).

\head{Model variants and metrics.} To demonstrate the effectiveness of our vision-language pre-training, we first include a baseline model without this pre-training. We then include two extreme settings of our model with $\lambda=1$ (\stos pre-training only) and $\lambda=0$ (\bidir pre-training only) to study how each objective individually works with different downstream tasks. Our full model conducts joint training on the two objectives. The fine-tuning procedure is performed the same regardless of the pre-training configurations.
Regarding evaluation metrics, we use standard language metrics for image captioning, including Bleu@4, METEOR, CIDEr, and SPICE and the official measurement on accuracy for VQA, 
over different answer types including Yes/No, Number, and Other.

\head{Comparisons against SotAs.} Results comparing our methods and SotA methods on the test set are in Tab.~\ref{tab:test_results}. We include state-of-the-art published works (upper part of Tab.~\ref{tab:test_results}), unpublished works that are currently in submission (middle part), and our methods (lower part).
All the image captioning methods are single models, with cross-entropy optimization only for a fair comparison.
Our full model (Unified \vlpabbrns) outperforms SotA methods on three out of four metrics on COCO, overall accuracy on VQA 2.0, and all four metrics on Flickr30k.
The improvements are particularly sound on Flickr30k, where we get \textit{5.1\% absolute gain} on CIDEr metric and \textit{2.8\%} on BLEU@4.

We further perform CIDEr optimization on COCO Captions through Self-Critical Sequence Training (SCST)~\cite{rennie2017self}, as in most of the recent image captioning literatures. The results are in Tab.~\ref{tab:cider_optim} where our full model sets new SotA on all the metrics.

\head{Boost from pre-training.} Our full model leads our baseline model by a large margin on most of the metrics thanks to our pre-training. Some noticeable improvements include over \textit{10\% absolute gain} on CIDEr metric on Flickr30k, and over \textit{2\% gain} on CIDEr on COCO and B@4, METEOR on Flickr30k. Small datasets (\ie, Flickr30k) benefit the most as vision-language pre-training alleviates overfitting issues.
Our model variants under the two extreme settings work well as expected on their ``favorable'' tasks, \ie, \stos pre-training alone improves downstream captioning tasks significantly and \bidir pre-training benefits understanding tasks (\ie, VQA), but not the opposite. They set new SotAs on all metrics except the ``Number'' accuracy on VQA 2.0. 
The joint training organically combines the representations learned from the two rather different objectives and yields slightly compromised but decent accuracy on all the downstream tasks.
That said, from an engineering perspective, if we can afford having separate pre-training models for generation task or understanding task, we will get the optimal model performance.
If we value model architecture and parameter sharing, the joint model is a good trade-off.

\begin{figure*}[t]
    \centering
    \includegraphics[width=\linewidth]{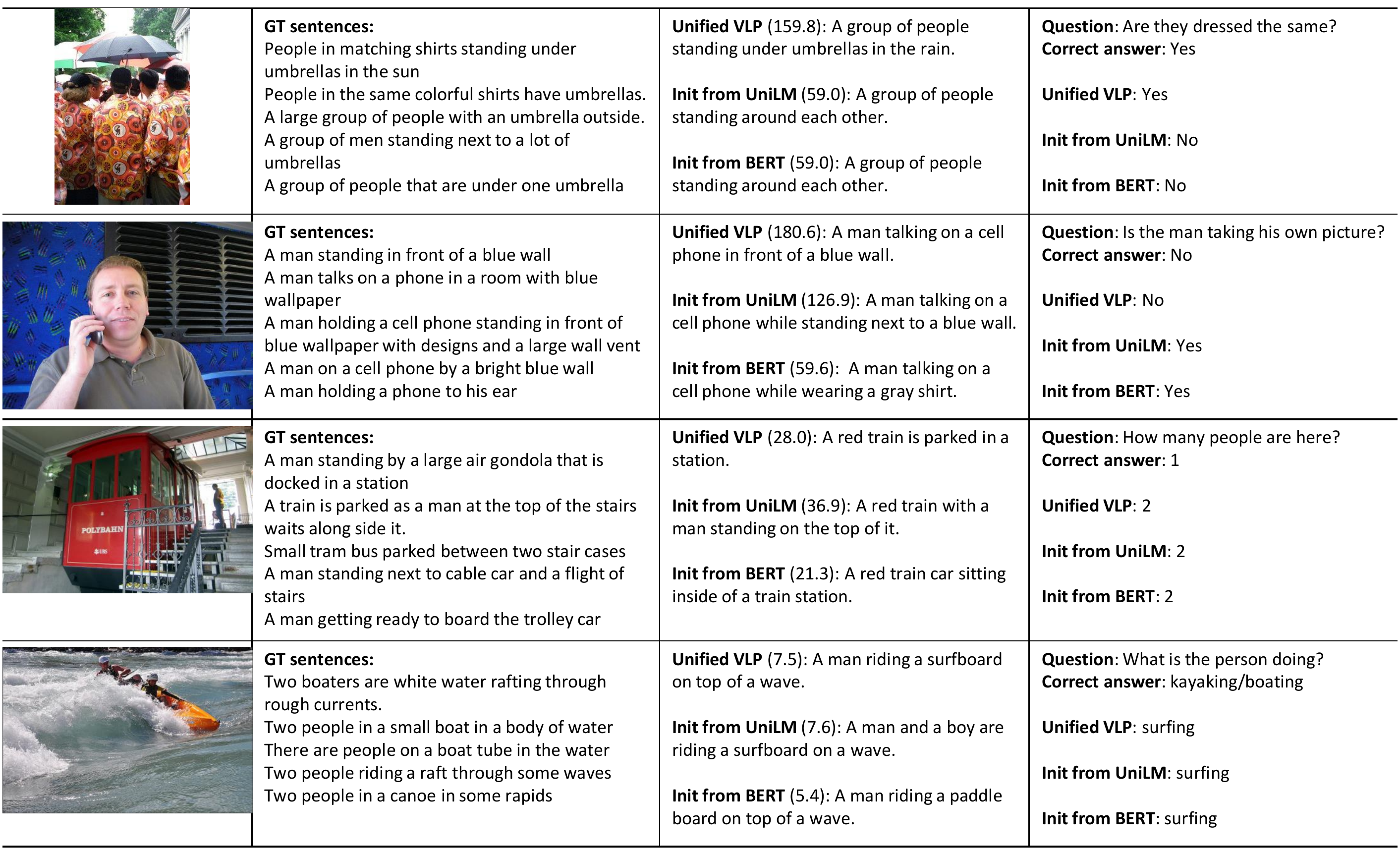}
    \caption{Qualitative examples on COCO Captions and VQA 2.0. The first column indicates images from the COCO validation set. The second column shows the five human-annotated ground-truth (GT) captions. The third column indicates captions generated by three of our methods and the corresponding CIDEr scores, where only Unified VLP has vision-language pre-training. The last column shows VQA questions and correct answers associated with the image and answers generated by our models. The top two are successful cases and the bottom two are failed cases. See text for details.}
    \label{fig:vis_examples}
\end{figure*}

\begin{table}[t]
    \centering
    {\small
    \begin{tabular}{l|rrrr}
        \toprule
        Method & B@4 & M & C & S  \\
        \midrule
        Region label as pretext & 5.4 & 9.4 & 62.2 & 14.5 \\ 
        Region label probability as input & 5.8 & 9.7 & 67.0 & 15.5  \\
        \bottomrule
    \end{tabular}
    }
    \caption{Comparison between having region class prediction pretext and feeding in class probabilities as a part of the model input. Results are on Conceptual Captions val set.} 
    \label{tab:vis_mask_results}
\end{table}

\head{Impact of pre-training types.}
Depending on how the base model Transformer is initialized, we define four ``degrees'' of pre-training from weakest to strongest as i) without any pre-training, \ie, base model is trained from scratch, ii) \bidir language pre-training, \ie, base model is initialized from BERT weights ~\cite{devlin2018bert}, iii) \stos and \bidir language pre-training, \ie, base model is initialized from UniLM weights~\cite{dong2019unified} which is our baseline setting, and iv) our full \vlpns.
The corresponding fine-tuning results on downstream tasks are presented in Fig.~\ref{fig:teaser} on the val set (full results see Appendix) and Tab.~\ref{tab:results_pt_levels} on the test set.
As shown from the figure, our vision-language pre-training significantly accelerates the learning process of downstream tasks and contributes to better overall accuracy. It is worth noting that the learning process of VQA is greatly shortened despite that the hidden states associated with tokens \cls and \sep are not learned during the pre-training. This indicates that the contextualized vision-language representations can generalize to unseen domains and work reasonable well as a warm-start for new tasks.

We also study how the pre-training types 1-3 influence our vision-language pre-training in terms of caption generation. The results on Conceptual Captions val set at epoch 20 are shown in  Tab.~\ref{tab:cc_results}.
All the models are trained based on the unified \vlpabbr objective ($\lambda=0.75$) for a fair comparison.
We observe that initializing base model with weights transferred from pure language pre-training benefits vision-language pre-training. The training objectives of UniLM are closer to our \stos and \bidir objectives than the ones in BERT and hence we hypothesize that this counts for the slightly larger improvement.
Note that our intention here is to demonstrate how different weight initializations can influence pre-training performance rather than pursuing possibly high quantitative scores (with full \stos training, CIDEr could climb to 77.2 after training for 30 epochs).

\head{Region object labels as pretext.}
Existing works~\cite{Zhou_2019_CVPR,lu2018neural} regard region object labels (probabilities) ($C_i$) as an important auxiliary to enrich image region features and here we follow a similar design.
We can also instead use these labels for a masked region classification pretext as in \cite{tan2019lxmert}. Here we have a comparison over the two design choices.
``region label probability as input'' is equivalent to our full model Unified \vlpabbr and ``region label as pretext'' is the implementation from~\cite{tan2019lxmert}. As shown in the results,
predicting class labels as a pretext has a negative impact on the pre-training, in terms of captioning performance.
We hypothesize that this is because the class labels from the off-the-shelf object detector might be noisy which compromises the learned feature representation. In contrast, our model refines the visual representation through a more reliable masked language modeling and could correct the errors exist in the class labels.

\head{Qualitative results and analyses.}
Qualitative examples on COCO Captions and VQA 2.0 are shown in Fig.~\ref{fig:vis_examples}. In the first two examples, our full model with vision-language pre-training captures more details in the image, such as ``umbrellas'' and ``a blue wall'' than the baseline methods. It also answers questions correctly. In the third example, all the methods dis-identify the gondola as a train due to their visual similarity. When it comes to the question answering, our methods all give correct answers while the GT answer is incorrect (note that there is a person in the gondola). In the fourth example, all the models mistakenly classify the activity as surfing while the correct one is kayaking/boating. This is consistent across both the caption model and the VQA model, which implies that the feature representations are indeed shared across tasks.

\begin{table*}[t]
    \centering
    {\small
    \begin{tabular}{l|rrrr|rrrr|rrrr} 
        \toprule
        & \multicolumn{4}{c}{COCO}  & \multicolumn{4}{c}{VQA 2.0} & \multicolumn{4}{c}{Flickr30k} \\
        Method & B@4 & M & C & S & Overall & Yes/No & Number & Other & B@4 & M & C & S \\
        \midrule
        From scratch & 34.5 & 28.1 & 114.2 & 21.1 & 63.4 & 80.2 & 46.4 & 55.2 & 26.9 & 20.8 & 52.1 & 14.4 \\
        Init from BERT & 34.6 & \textbf{28.4} & 114.8 & 21.4 & 65.1 & 82.9 & 48.0 & 56.1 & 27.5 & 21.9 & 58.4 & 15.5  \\
        \midrule
        Init from UniLM \\
        \;\;w/o \vlpabbr pre-training (baseline) & 34.5 & 28.1 & 113.9 & 21.3 & 66.1 & 83.8 & 49.7 & 56.9 & 27.5 & 21.5 & 58.3 & 15.3 \\
        \;\;\stos pre-training only & \textbf{35.3} & \textbf{28.4} & \textbf{116.7} & \textbf{21.5} & 66.4 & 84.6 & \textbf{50.1} & 56.9 & 28.9 & \textbf{23.6} & 67.0 & \textbf{17.2}  \\
        \;\;\bidir pre-training only & \textbf{35.3} & 28.3 & 116.1 & 21.4 & \textbf{68.2} & \textbf{85.6} & \textbf{51.9} & \textbf{59.3} & \textbf{29.6} & 23.2 & \textbf{67.2} & 16.8 \\
        \;\;Unified \vlpabbr & \textbf{35.5} & \textbf{28.5} & \textbf{118.0} & \textbf{21.6} & \textbf{67.4} & \textbf{85.4} & \textbf{50.1} & \textbf{58.3} & \textbf{29.7} & \textbf{23.8} & \textbf{69.1} & \textbf{17.6}  \\
        \bottomrule
    \end{tabular}
    \caption{Results on COCO Captions, VQA 2.0, and Flickr30k validation set. B@4 represents for BLEU@4, M for METEOR, C for CIDEr, and S for SPICE. Top two results on each metric are in bold.}
    \label{tab:val_results}
    }
\end{table*}

\begin{table*}[t]
    \centering
    \small{
    \begin{tabular}{l|rrrrr}
        \toprule
        Dataset & Batch Size & Learning Rate & \# of Epochs & GPUs & Time per Epoch \\
        \midrule
        CC & 64(x8) & 1e-4(x8) & 30 & 8x V100 & 5hr \\ 
        \midrule
        COCO & 64(x8) & 3e-5(x8) & 30 & 8x V100 & 12min \\ 
        VQA 2.0 & 64(x2) & 2e-5(x2) & 20 & 2x V100 & 32min \\ 
        Flickr30k & 64(x8) & 3e-5(x8) & 30 & 8x V100 & 3min \\ 
        \midrule
        COCO (w/o pre-training) & 64(x8) & 3e-4(x8) & 30 & 8x V100 & 12min \\ 
        COCO (SCST training) & 16(x4) & 1e-6(x4) & 30 & 4x Titan Xp & 3hr \\ 
        \bottomrule
    \end{tabular}
    }
    \caption{Model hyper-parameters and training specifications.} 
    \label{tab:hyper-param}
\end{table*}

\section{Conclusion}
This paper presents a unified Vision-Language Pre-training (VLP) model that can be fine-tuned for both vision-language generation and understanding tasks.
The model is pre-trained on large amounts of image-text pairs based on two objectives: bidirectional and seq2seq vision-language prediction. The two disparate objectives are fulfilled under the same architecture with parameter sharing, avoiding the necessity of having separate pre-trained models for different types of downstream tasks (\ie, generation-based or understanding-based).
In our comprehensive experiments on image captioning and VQA tasks, we demonstrate that the large-scale unsupervised pre-training can significantly speed up the learning on downstream tasks and improve model accuracy.
Besides, compared to having separate pre-trained models, our unified model combines the representations learned from different objectives and yields slightly compromised but decent (SotA) accuracy on all the downstream tasks.
In our future work, we would like to apply VLP to more downstream tasks, such as text-image grounding and visual dialogue. Methodology-wise, we would want to see how multi-task fine-tuning can be applied to our framework to alleviate interference between different objectives.

\head{Acknowledgement.} The technical work was performed
during Luowei's summer internship at Microsoft Research. Luowei Zhou and Jason Corso were partly supported by DARPA FA8750-17-2-0125 and NSF IIS
1522904 as part of their affiliation with University of Michigan. This article solely reflects the opinions and conclusions of its authors but not the DARPA or NSF. We thank Li Dong and Furu Wei for generously sharing us their UniLM source code. We thank Kezhen Chen for his helpful discussions.

\newpage
\section{Appendix}
\subsection{Results on Downstream Tasks}
We include the validation results on fine-tuning tasks in Tab.~\ref{tab:val_results}. Note that for VQA 2.0, all the methods here are only trained on the training set while for the results reported on the test set (Tab. 3 and Tab. 4 in the main paper), all the models are trained on both training set and validation set following the practice from early works.

\subsection{Implementation Details}
\label{sec:implementation:details}
\head{Region proposal and feature.} We use a variant of Faster RCNN model~\cite{ren2015faster} with ResNeXt-101 FPN backbone~\cite{xie2017aggregated} for region proposal and feature extraction. The Faster RCNN model is pre-trained on the Visual Genome dataset~\cite{krishna2017visual}, following the same procedure in~\cite{anderson2018bottom} for joint object detection (1600 classes) and attribute classification.
We set the number of regions per image to exact 100 as suggested in~\cite{jiang2018pythia}. We take the output of the fc6 layer as the feature representation for each region, and fine-tune the fc7 layer.

\head{Model hyper-parameters.} The model hyper-parameters on pre-training and fine-tuning are in Tab.~\ref{tab:hyper-param}. The SCST training on COCO is performed after the \vlpabbr pre-training and COCO fine-tuning.

\head{Training details.}
We use the same training optimizer as in BERT~\cite{devlin2018bert} and other training  hyper-parameters are in Tab.~\ref{tab:hyper-param}. Our VQA models are trained on 2x V100 GPUs, COCO Captions SCST training on 4x Titan Xp GPUs, and all others are on 8x V100 GPUs. \\

\newpage
\begin{small}
\bibliographystyle{aaai}
\bibliography{AAAI-ZhouL.362}
\end{small}

\end{document}